\documentclass{ecai}

\usepackage{latexsym}
\usepackage{amssymb}
\usepackage{amsmath}
\usepackage{amsthm}
\usepackage{booktabs}
\usepackage{enumitem}
\usepackage{graphicx}
\usepackage{color}
\usepackage{hyperref}
\usepackage{url}
\usepackage{algorithm}
\usepackage{algorithmic}
\usepackage{graphicx}
\usepackage{booktabs}
\usepackage{xcolor}
\newcommand{\BibTeX}{B\kern-.05em{\sc i\kern-.025em b}\kern-.08em\TeX}

\begin{document}


\begin{frontmatter}

\title{Accelerating Transformers in Online RL}

\author[A]{\fnms{Daniil}~\snm{Zelezetsky}\thanks{Corresponding Author. Email: zelezetsky@iaipht.ru.}}
\author[B, A]{\fnms{Alexey K.}~\snm{Kovalev}}
\author[B, A]{\fnms{Aleksandr I.}~\snm{Panov}} 

\address[A]{IAI MIPT, Moscow, Russia}
\address[B]{Cognitive AI Lab, Moscow, Russia}

\begin{abstract}

The appearance of transformer-based models in Reinforcement Learning (RL) has expanded the horizons of possibilities in robotics tasks, but it has simultaneously brought a wide range of challenges during its implementation, especially in model-free online RL. Some of the existing learning algorithms cannot be easily implemented with transformer-based models due to the instability of the latter. In this paper, we propose a method that uses the Accelerator policy as a transformer's trainer. The Accelerator, a simpler and more stable model, interacts with the environment independently while simultaneously training the transformer through behavior cloning during the first stage of the proposed algorithm. In the second stage, the pretrained transformer starts to interact with the environment in a fully online setting. As a result, this model-free algorithm accelerates the transformer in terms of its performance and helps it to train online in a more stable and faster way. By conducting experiments on both state-based and image-based ManiSkill environments, as well as on MuJoCo tasks in MDP and POMDP settings, we show that applying our algorithm not only enables stable training of transformers but also reduces training time on image-based environments by up to a factor of two. Moreover, it decreases the required replay buffer size in off-policy methods to 10–20 thousand, which significantly lowers the overall computational demands.

The code is available at: \href{https://github.com/Dzelezetsky/Accelerating_Transformers}{\nolinkurl{github.com/Dzelezetsky/Accelerating_Transformers}}

\end{abstract}

\end{frontmatter}


\section{Introduction}
Transformers~\citep{NIPS2017_3f5ee243} have demonstrated remarkable success in various domains due to their ability to model long-range dependencies and complex patterns~\cite{bulatov2022recurrent,cherepanov_recurrent_2024}. In the field of robotics and reinforcement learning (RL), one of the main advantages of transformers is their suitability for processing multi-modal data such as text and images~\citep{vima, openvla, radford21a, gato, octo}.
There are three main approaches to train transformers in RL: 1) fully offline training~\citep{qtrans, chen2021decision,traj_transformer,mgdt,strap}, 2) hybrid methods with both offline and online training~\citep{retrieval, RL+Transformer, sun2023smart, online_dt}, and 3) fully online training without offline data~\citep{dtqn, top_erl, gtrxl,relit,ctsac}.

Offline RL~\citep{chen2021decision, traj_transformer} is a widely used approach for training transformer models. It is beneficial to use offline data to lower training costs, which is particularly important in fields such as robotics or autonomous transport. Despite its advantages, offline RL has several limitations that hinder agent performance and create barriers to its widespread adoption~\citep{should}. One of the main weaknesses is the effort required to collect expert demonstrations. Sophisticated environments often require large models which, in turn, demand a significant amount of training data to perform well. Moreover, to maintain high performance during evaluation or real-world deployment, attention must be paid to distributional shifts between offline trajectories and the real state distribution. While online RL addresses this issue through extensive interaction with the environment and by tuning the exploration coefficient, offline training does not provide such flexibility. These circumstances can potentially limit agent performance by weak generalization and underscore the need for a large and high-quality dataset. 

Online training of transformers can theoretically address such drawbacks of offline training as the need to collect a large dataset, state distribution shifts, and lack of exploration. However, in practice, online training of transformers tends to be less stable, which limits the applicability of online algorithms for this architecture~\citep{gtrxl}. Transformers are known to be sensitive to hyperparameters and optimization settings, which makes their training process inherently unstable~\citep{gtrxl}. In online RL, where the agent interacts with the environment in real-time, this instability is also exacerbated by the non-stationary nature of the data distribution. 
Transformers typically require large amounts of data to achieve good performance, so in online RL, where data is collected incrementally through interactions with the environment, this is time-consuming. 

In this paper, we propose an algorithm that accelerates (pretrains) the transformer in an online manner. During the first stage, the accelerator policy generates trajectories used to pretrain the transformer. In the second stage, the pretrained transformer interacts with the environment on its own in a fully online setting. The choice of the accelerator's architecture depends on the nature of the task: whether the environment can be described by a Markov Decision Process (MDP) or a Partially Observable Markov Decision Process (POMDP), or whether we process a vector-based environment or an image-based one.

Our proposed approach addresses problems such as training instability, sample inefficiency, and exploration limitations, and avoids the need for offline data. Despite the fact that the first stage of our method utilizes BC from the Accelerator's demonstrations, it still maintains the ability to explore the environment by tuning Accelerator's exploration during the collection of training data. By pretraining the transformer on these demonstrations, we also tackle sample inefficiency. Our method avoids environment distribution shifts by letting the transformer explore the environment by itself in the second stage of the training procedure. 
We also conduct experiments on MDP and POMDP vector-based robotic locomotion~\citep{mujoco} and manipulation~\citep{maniskill3} tasks, and compare accelerated transformer with MLP and LSTM~\citep{lstm} baselines.

Our contributions are as follows:
\begin{enumerate}
    \item We propose the ``Accelerator" — a flexible and easy-to-tune training framework that pretrains a transformer in an online manner, eliminating the need for an offline dataset.
    \item Through experiments, we empirically demonstrate the effectiveness of the proposed algorithm on robotic locomotion and manipulation tasks. We also show that the Accelerator can reduce training time by up to a factor of two and decrease the required replay buffer size for the transformer to as little as 10–20 thousand, which substantially lowers computational resource demands.
\end{enumerate}

\section{Related Work}
There is a growing body of work that utilizes transformers in RL, adapting their architecture and training algorithms to ensure stable performance.
\citet{chen2021decision} proposed the Decision Transformer (DT), which reduces the RL task to supervised sequence modeling and serves as a popular baseline for transformer-based offline RL. Thus, DT-like architectures can be adapted for multi-task learning~\citep{mgdt}, or for more flexible approaches with beam-search action generation~\citep{traj_transformer}. However, offline data can consist of sub-optimal trajectories, so sequence modeling approaches could affect policy performance. To address this problem, Q-learning Decision Transformer~\citep{qdt} leverages dynamical programming (especially Q-learning) and relabel return-to-go in order to train DT on new data. Another problem with offline RL is the limited amount of training data. To overcome this problem,~\citet{bootstrap} proposed the Bootstrapped Transformer which relies on bootstrapping ideas and generates a synthetic dataset for the trained model. 

Offline pretraining with further online fine-tuning eliminates some of the problems of the previous approach, especially since it can regulate exploration during online interaction with the environment~\cite{staroverov_fine-tuning_2023}. SMART algorithm~\citep{sun2023smart} separates the training process into two steps: 1) Offline self-supervised sequence modeling and 2) Online fine-tuning. \citet{awac} proposed AWAC, an algorithm that enables rapid fine-tuning with a combination of prior demonstration data and online experience. \citet{offline2online} proposed a training algorithm that can learn from a small amount of demonstrations, while classical Behavior Cloning (BC) requires more data to achieve the same result. The authors of the Online Decision Transformer~\citep{online_dt} use an offline dataset and minimize the log-likelihood of expert demonstrations in order to stabilize the agent during online training. This technique uses pre-collected data simultaneously with online training.

Fully online training remains a complex and underexplored problem in transformer-based RL~\citep{kachaev2025a}. Building on the Transformer-XL (TrXL) architecture~\citep{trxl}, Gated Transformer-XL (GTrXL)\citep{gtrxl} introduces gating mechanisms that regulate skip-connections, enabling effective learning in memory-demanding tasks\citep{pleines2023memory}. \citet{dtqn} proposed Deep Transformer Q-Network, adapting transformers to DQN-based training.

Transformers’ sequence modeling ability makes them well-suited for Episodic RL (ERL). \citet{top_erl} extend transformers to generate action sequences and train them with a tailored off-policy algorithm. Their TOP-ERL method significantly outperforms GTrXL and leading RL algorithms such as PPO and SAC. Transformers have also been integrated into model-based RL: \citet{qt-tdm} introduced QT-TDM, which predicts environment dynamics in an online setting.

Beyond direct interaction, transformers have been trained via knowledge distillation. \citet{ald} applied actor-learner distillation to improve sample efficiency and reduce training time. However, these methods are mostly limited to discrete action spaces.

Another line of work adapts RL training across architectures. Policy-Agnostic RL (PA-RL)~\citep{pa-rl} presents a unified actor-critic framework for offline RL and online fine-tuning that supports diverse policy classes, including transformers. By decoupling policy improvement from parameterization through supervised learning on optimized actions, PA-RL achieves state-of-the-art results and demonstrates the first successful fine-tuning of large pre-trained robotic transformer policies such as OpenVLA in real-world settings.

The objective of our algorithm is not to introduce a radically new adaptation of transformers for model-free online RL, but to propose a stable framework that unifies the full training cycle of a transformer. The proposed framework combines the advantages of offline methods, such as stable and efficient learning from expert trajectories, with those of online approaches, including the ability to regulate exploration and the amount of environment interaction, which improves robustness to state distribution shifts and enhances generalization.

\section{Reinforcement Learning}

Markov Decision Process (MDP) is defined as a tuple: $\mathcal{M} = (\mathcal{S}, \mathcal{A}, \mathcal{P}, \mathcal{R}, \gamma)$ where: $\mathcal{S}$ is a set of states, $\mathcal{A}$ is a set of actions, $\mathcal{P}(s' | s, a)$ is a state transition probability function, $\mathcal{R}(s, a, s')$ -- a reward function, and $\gamma \in [0, 1]$ -- a discount factor.
The goal in the RL is to find a policy \(\pi^*(a | s)\) that maximizes the expected cumulative reward $\mathbb{E}\left[ \sum_{t=0}^\infty \gamma^t r_t \right]$ where \(r_t\) is the reward received at time \(t\).
In the partially observable setting, the environment is modeled as a Partially Observable Markov Decision Process (POMDP), defined by a tuple $\mathcal{M}_{\text{POMDP}} = (\mathcal{S}, \mathcal{A}, \mathcal{P}, \mathcal{R}, \Omega, \mathcal{O}, \gamma)$, where $\Omega$ is a set of observations and $\mathcal{O}(o \mid s)$ is the observation probability function. The agent selects actions based on a belief distribution \(b(s)\) over latent states, and the optimal policy \(\pi^*(a \mid b)\) maximizes $\mathbb{E}_{b_0, \pi} \left[ \sum_{t=0}^\infty \gamma^t r_t \right]$.

\section{Proposed Method}

To accelerate transformer training, we employ an accelerator that generates training data for the first stage. The transformer is then finetuned in the environment on its own during the second stage. This two-stage process provides both stable pretraining for foundational skills and online finetuning for precise task adaptation. Although this section provides an explanation based on an \textbf{MLP-based accelerator}, any other model design choices are still compatible and can be used instead (\autoref{sec:experiments1} demonstrates results with LSTM-based accelerator in POMDP environments). The main rule is the following: the accelerator may have weaker performance in the environment, but it must be more stable in terms of training. This ensures that the accelerator will bring the transformer to a level sufficient for further online fine-tuning. Although we explain our approach using Twin Delayed Deep Deterministic Policy Gradient~\citep{td3} (TD3) algorithm, it can be generalized to other off-policy actor-critic algorithms such as Deep Deterministic Policy Gradient~\citep{ddpg} or Soft Actor-Critic~\citep{sac} (SAC) (the fourth research question in~\autoref{sec:experiments1} discusses the experimental results obtained with the SAC algorithm). To achieve optimal performance of the transformer, the second stage of our method releases it into fully independent fine-tuning in the online setting. Having completed the first stage, the transformer is sufficiently pre-trained to stably strengthen its ability to interact with the environment.

\subsection{The First Stage: Transformer Acceleration} \label{dubsec:one} In the first stage, the accelerator policy trains by itself while simultaneously serving as a trajectory generator for the transformer policy. Let us define the accelerator policy as $\pi_{\phi}(a|s)$ and the transformer policy as $\pi_{\theta}(a|s)$. During the first stage, $\pi_{\phi}(a|s)$ trains via the standard RL pipeline, with the exception that it saves states and actions in a special trajectory buffer $\mathcal{T}$, which is then used as a dataset to train the transformer.

\begin{algorithm}[tb]
   \caption{TD3-based \textbf{acceleration} stage algorithm}
   \label{alg:first_stage}
\begin{algorithmic}[1] 
   \STATE Initialize accelerator's critics $Q_{\phi_1}$, $Q_{\phi_2}$, actor $\pi_\phi$, and \textcolor{red}{transformer actor $\pi_\theta$}
   \STATE Initialize target networks $\phi'_1 \leftarrow \phi_1$, $\phi'_2 \leftarrow \phi_2$, $\phi' \leftarrow \phi$, \textcolor{red}{$\theta' \leftarrow \theta$}
   
   \STATE Initialize replay buffer $\mathcal{B}$ and \textcolor{red}{trajectory buffer $\mathcal{T}$}
   \FOR{$t=1$ {\bfseries to} $T$}
   \STATE Select action with exploration noise $a \sim \pi_\phi(s) + \epsilon$, 
   \STATE $\epsilon \sim \mathcal{N}(0, \sigma)$ and observe reward $r$ and new state $s'$
   \STATE Store transition tuple $(s, a, r, s')$ in $\mathcal{B}$
   \IF{$t > C$}
      \STATE \textcolor{red}{Store tuple $(\{s_{t-C+1},...,s_t\}, a)$ in $\mathcal{T}$}
   \ENDIF   
   \STATE Sample mini-batch of $N$ transitions $(s, a, r, s')$ from $\mathcal{B}$  
   \STATE $\tilde a \leftarrow \pi_{\phi'}(s') + \epsilon, \quad \epsilon \sim clip(\mathcal{N}(0, \tilde \sigma), -c, c)$
   \STATE $y \leftarrow r + \gamma \min_{i=1,2} Q_{\phi'_i}(s', \tilde a)$
    \STATE Update critics $\phi_i \leftarrow argmin_{\phi_i} N^{-1} \sum (y - Q_{\phi_i}(s,a))^2$
   \IF{time to update} 
   \STATE \textcolor{red}{Train transformer by~\autoref{alg:alg3} and optionally~\autoref{alg:alg4} then clear $\mathcal{T}=\emptyset$} 
   \ENDIF 
   \IF{$t$ mod $d$}
   \STATE Update $\phi$ by the deterministic policy gradient:
   \STATE $\nabla_{\phi} J(\phi) = N^{-1} \sum \nabla_{a} Q_{\phi_1}(s, a) |_{a=\pi_{\phi}(s)} \nabla_{\phi} \pi_\phi(s)$
   \STATE Update target networks:
   \STATE $\phi'_i \leftarrow \tau \phi_i + (1 - \tau) \phi'_i$
   \STATE $\phi' \leftarrow \tau \phi + (1 - \tau) \phi'$
   \ENDIF
   \ENDFOR
\end{algorithmic}
\end{algorithm}

Algorithm \autoref{alg:first_stage} describes the process of accelerator training and transformer acceleration based on the TD3 training pipeline. All changes made to the standard TD3 algorithm are highlighted in red. In particular, rows 5, 6, 7, and 12 clarify how trajectories are collected in $\mathcal{T}$ and used for transformer acceleration: the sequence of states is accumulated until the number of Accelerator steps from the beginning of the episode exceeds the pre-selected transformer context size $C$. Once this occurs, the algorithm starts storing the training data into $\mathcal{T}$. Note that this approach can also be applied during the accelerator’s evaluation process.

During the transformer's update phase, it can utilize either behavior cloning alone or do it with additional gradient ascent over accelerator's critic function $Q_{\phi_i}(s,a)$, as described in Algorithm~\ref{alg:alg3} and Algorithm~\ref{alg:alg4}, respectively. This is considered an optional enhancement that may improve acceleration but is not strictly necessary. Since the accelerator is trained by actor-critic algorithm, we want its actor $\pi_\phi$ to ascend its critic’s function, which approximates the state-action function $Q_{\phi_i}(s,a)$. Therefore, it becomes possible to improve transformer training by adjusting its weights in the direction of the critic’s gradient ascent, as in Algorithm~\ref{alg:alg4} (a more detailed discussion of this gradient ascent is provided later in~\autoref{sec:experiments1}).

\begin{algorithm}[tb]
\caption{Behavior cloning}
\label{alg:alg3}
\begin{algorithmic}[1]
\REQUIRE Transformer $\pi_{\theta}(a|s)$,  $\mathcal{T}$
\FOR{$(\{s_{t-C+1},...,s_t\}, a)$ in $\mathcal{T}$}
    \STATE Make prediction $\hat{a}=\pi_{\theta}(s_{t-C+1},...,s_t)$
    \STATE Calculate $L = MSE(a,\hat{a})$
    \STATE Update actor $\theta_{t+1} = \theta_t - \alpha \nabla_{\theta} L$
    \STATE Update target $\theta'_{t+1} \leftarrow \tau \theta_{t+1} + (1 - \tau) \theta'_t$
\ENDFOR
\end{algorithmic}
\end{algorithm}

\begin{algorithm}[tb]
\caption{Ascending on critic}
\label{alg:alg4}
\begin{algorithmic}[1]
\REQUIRE Transformer $\pi_{\theta}(a|s)$, dataset $\mathcal{T}$, critic $Q_{\phi_1}$
\FOR{$(\{s_{t-C+1},...,s_t\}, a)$ in $\mathcal{T}$}
    \STATE Make prediction $\hat{a} = \pi_{\theta}(s_{t-C+1},...,s_t)$
    \STATE Evaluate $Q_{\phi_1}(s, \hat{a})$ using the accelerator's critic
    \STATE Update $\theta_{t+1} = \theta_t + \alpha \nabla_{\hat{a}} Q(s, \hat{a}) \nabla_{\theta} \hat{a}$
    \STATE Update target $\theta'_{t+1} \leftarrow \tau \theta_{t+1} + (1 - \tau) \theta'_t$
\ENDFOR
\end{algorithmic}
\end{algorithm}


During the first stage, the transformer policy forms its basic structure of weights that will assist it in online fine-tuning which is the second stage of our proposed algorithm. This approach leverages supervised learning, which is beneficial in terms of the transformer's stability. Notably, since we generate our transformer’s training data in a real-time manner, we can regulate its quality and diversity by dynamically changing the parameters of the Accelerator’s training. For instance, to increase exploration, we can add Gaussian noise to the states from the environment, $\tilde s = s + \mathcal{N}(0,\sigma)$, sample target actions according to these observations, $a \sim \pi_\phi(\tilde s)$, store $(\tilde s, a)$ in $\mathcal{T}$, and use this data within Algorithm~\ref{alg:alg3}. In addition, the length of the Accelerator's training session can be as long as needed, so the amount of training data can be adjusted to match the task’s requirements.
These measures help overcome well-known offline RL challenges such as the exploration problem, dataset size limitations, and state distribution shifts. All the architecture and training details are available in~\autoref{appendix}.

The trigger for completing the first stage can be either reaching a predefined performance threshold of the transformer, hitting a limit on environment steps or training time, manual termination once the required performance is achieved, or automatic stopping when performance plateaus, indicating that the transformer has ceased progressing during the pretraining phase.

\subsection{The Second Stage: Online Fine-Tuning}
The first stage yields a pretrained transformer policy which is ready to continue its training in the fully online setting. At this stage, we can drop the accelerator’s actor but continue using its critic in order to conjugate it with the transformer actor. Starting from this time, these two models operate together in a standard online RL training pipeline, described in Algorithm~\ref{alg:second_stage}.

\begin{algorithm}[!t]
   \caption{TD3-based \textbf{fine-tuning} stage algorithm}
   \label{alg:second_stage}
\begin{algorithmic}[1]   
   \STATE Use $Q_{\phi_1}$, $Q_{\phi_2}$,$\pi_\theta$ with their targets from the first stage, initialize empty replay buffer $\mathcal{B}$
   \FOR{$t=1$ {\bfseries to} $T$}
   \STATE Select action with exploration noise $a \sim \pi_\theta(s) + \epsilon$, 
     and observe reward $r$ and new state $s'$
   \STATE Store transition tuple $(s, a, r, s')$ in $\mathcal{B}$
   \STATE Sample mini-batch of $N$ transitions $(s, a, r, s')$ from $\mathcal{B}$  
   \STATE $\tilde a \leftarrow \pi_{\theta'}(s') + \epsilon, \quad \epsilon \sim clip(\mathcal{N}(0, \tilde \sigma), -c, c)$
   \STATE $y \leftarrow r + \gamma \min_{i=1,2} Q_{\phi'_i}(s', \tilde a)$
    \STATE Update critics $\phi_i \leftarrow \arg\min_{\phi_i} \; N^{-1} \sum (y - Q_{\phi_i}(s,a))^2$

   \IF{$t$ mod $d$}
   \STATE Update $\theta$ by the deterministic policy gradient:
   \STATE $\nabla_{\theta} J(\theta) = N^{-1} \sum \nabla_{a} Q_{\phi_1}(s, a) |_{a=\pi_{\theta}(s)} \nabla_{\theta} \pi_\theta(s)$
   \STATE Update target networks:
   \STATE $\phi'_i \leftarrow \tau \phi_i + (1 - \tau) \phi'_i$
   \STATE $\theta' \leftarrow \tau \theta + (1 - \tau) \theta'$
   \ENDIF
   \ENDFOR
\end{algorithmic}
\end{algorithm}

Algorithm~\ref{alg:second_stage} describes the fine-tuning stage of the transformer. As mentioned earlier, the acceleration stage returns three networks: the accelerator actor $\pi_{\phi}(a|s)$, the accelerator critic $Q_{\phi_i}(i=1,2)$, and the transformer $\pi_{\theta}(a|s)$. During the fine-tuning stage, we continue to use the accelerator critic alongside the transformer for further online training. In our TD3-based algorithm description, the fine-tuning stage follows a standard training process using the same TD3 algorithm, where the actor is the transformer and the critic is the accelerator critic.

As we previously noted, the advantage of our algorithm lies in its complete flexibility when choosing architectures for the accelerator. This means that in a POMDP setting, it is possible to select a critic architecture capable of processing state sequences, thereby preventing the critic’s architectural limitations from affecting the transformer's fine-tuning stage.

\section{Research Questions}
\label{sec:exp}

 This section formulates the research questions (RQs) that need to be addressed in order to fully assess the usefulness of the algorithm. Answers to these questions will help clarify the feasibility of applying this algorithm, its advantages compared to training a transformer from scratch without the acceleration phase, and provide insights into the nuances of tuning the training process to achieve maximum effectiveness. Within this section, we have identified five main questions:
\begin{enumerate}
    \item Can transformer train well on the first stage?
    \item Can transformer achieve performance comparable to MLP and LSTM baselines?
    \item  Can additional gradient ascent (Algorithm~\ref{alg:alg4}) improve acceleration?
    \item Can ``Accelerator" be applied to other off-policy algorithms and image-based environments?
    \item Can ``Accelerator" improve training speed and reduce computational costs?
\end{enumerate}

\subsection{Environments} We evaluate the proposed method on two types of vector-based environments: MuJoCo~\citep{mujoco} (HalfCheetah, Ant, and Hopper) and ManiSkill~\citep{maniskill3} (PushCube, PullCube). We additionally employ the POMDP versions of the MuJoCo HalfCheetah, Ant, and Hopper environments (with masked position information) to evaluate the performance of the LSTM accelerator, which leverages sequential data and is better suited for non-Markovian tasks. To check if our method works on image-based tasks, we utilize RGB versions of ManiSkill PushCube and PullCube tasks.
MuJoCo is a physics engine designed for research and development in robotics. It is used to simulate locomotion tasks, where agents learn to move efficiently in complex environments. Our choice of this set of environments is motivated by their popularity for benchmarking models and algorithms, ease of implementation, and fast runtime.
ManiSkill is a simulation environment designed for robotic manipulation tasks, focusing on dexterous manipulation, object interaction, and task-oriented learning. We use ManiSkill due to its efficient and convenient parallelization of environments, low computational requirements, and the availability of both state-vector and image-based versions of the same task.

\subsection{Baselines}
In this paper, we use MLP, LSTM~\citep{lstm}, and Vanilla Transformer~\citep{NIPS2017_3f5ee243} baselines trained from scratch 
and compare the accelerated transformer's performance with the performance of the baselines. We use the MLP because of its ease of learning and stability; it can achieve high episodic rewards on both the MuJoCo and ManiSkill tasks we use. If we operate in an image-based environment, we utilize CNN to process images. Similar to the transformer, the LSTM architecture processes a sequence as input, but it is often easier to train in RL settings and is more robust to noisy or irregularly sampled data. These considerations motivate comparing the transformer with LSTM to see whether our algorithm allows us to train the transformer to achieve results comparable to those of the LSTM. Finally, comparing our method with an online-trained Vanilla Transformer highlights its advantages in terms of data efficiency and faster convergence. It is important to emphasize that our goal is not to design a state-of-the-art transformer modification but rather to propose a method for stabilizing their training. Therefore, when comparing with baselines, our primary objective is to demonstrate that the model can be trained reliably and achieve performance that is at least on par with them.

\subsection{Details}
All the charts in this section describe the training progress of the models in terms of evaluation reward (for MuJoCo environments) or evaluation success rate (for ManiSkill environments). Each progress curve also includes a standard deviation, obtained by averaging the key metric over 30 seeds with 1 parallel environment for MuJoCo tasks, and using a single seed and 50 parallel environments for ManiSkill tasks.
All experiments are conducted on a Tesla V100 GPU for vector environments and A100 for image environments, with 128GB of RAM. Additional information about the accelerators, transformers, and baseline models’ parameters, as well as their corresponding training parameters, is available in~\autoref{appendix}. The main paper highlights the most visually illustrative results of the experiments. Additional results, including those in other environments, are included in the corresponding appendix sections.

\subsection{Experiments}
\label{sec:experiments1}
\paragraph{RQ1. Can transformer train well on the first stage}

\begin{figure}[H]
    \centering
\includegraphics[width=\linewidth]{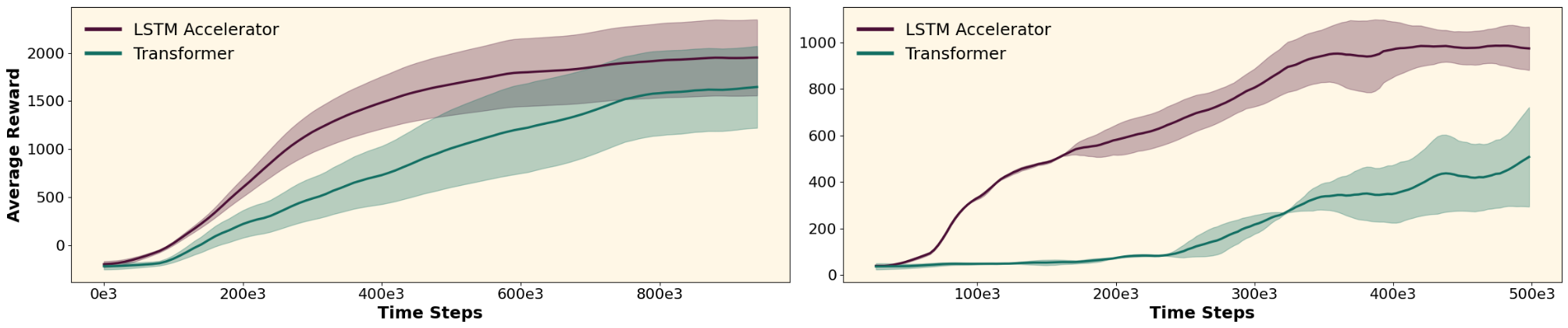}
    \caption{The first stage (transformer acceleration) in POMDP HalfCheetah (left) and POMDP Hopper (right) tasks. LSTM accelerator trains in the environment by itself using TD3 and simultaneously generating trajectories for transformer.}
    \label{fig:st1_main}
\end{figure}

The first question that arises when implementing our algorithm is whether the second stage (RL finetuning) is necessary. One might assume that since we can fully control the diversity of training data through critic-guided exploration, the transformer could potentially learn solely from this stage. To examine this, we present the training progress of the transformer during the first stage along with its accelerator. According to~\autoref{fig:st1_main}, on a broader scale, one can observe a clear tendency: the reward curve of the transformer follows the behavior of the accelerator’s curve. This is a natural outcome, as the transformer’s performance improves in line with the increasing returns of the accelerator. However, it is clearly visible that despite this, the transformer is unable to reach the performance level of its accelerator. Therefore, this does not prevent the transformer from further fine-tuning.
  
Same experiments in other environments can be found in~\autoref{sec: suplement_rq1}.

\paragraph{RQ2. Can transformer achieve performance comparable to MLP and
LSTM baselines?}

\begin{figure}[!t]
    \centering
    \includegraphics[width=\linewidth]{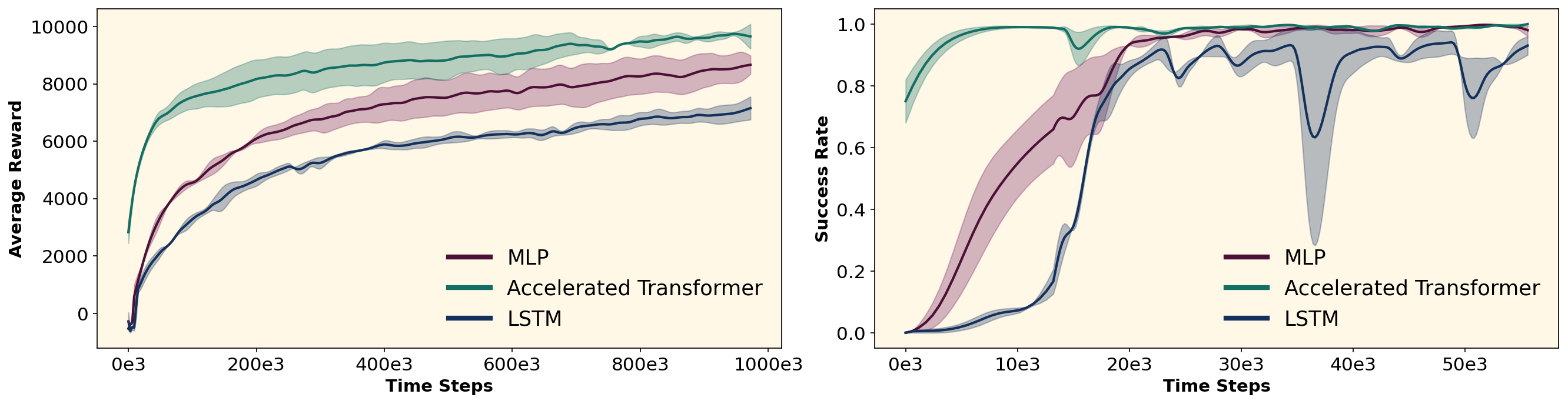}
    \caption{Average reward in HalfCheetah (left) and success rate in PushCube (right) tasks. Accelerated (pretrained) transformer demonstrates strong capabilities to finetune on both tasks.}
    \label{fig:stage2_main}
    \vspace{15pt}
\end{figure}

In this section, we evaluate the potential of the proposed algorithm and address whether the accelerated transformer can achieve performance comparable to MLP and LSTM. In this experiments, all the models utilize TD3 training algorithm. \autoref{fig:stage2_main} shows that the accelerated transformer reaches a performance competitive with MLP and LSTM in fewer training steps. According to the results in the HalfCheetah task (\autoref{fig:stage2_main}, left), the transformer can be fine-tuned to surpass the performance of MLP and LSTM. Results in the PushCube task (\autoref{fig:stage2_main}, right) demonstrate that it also requires significantly fewer steps to converge and achieve the maximum success rate. 

\begin{figure*}[t]
    \centering
    \includegraphics[width=\linewidth]{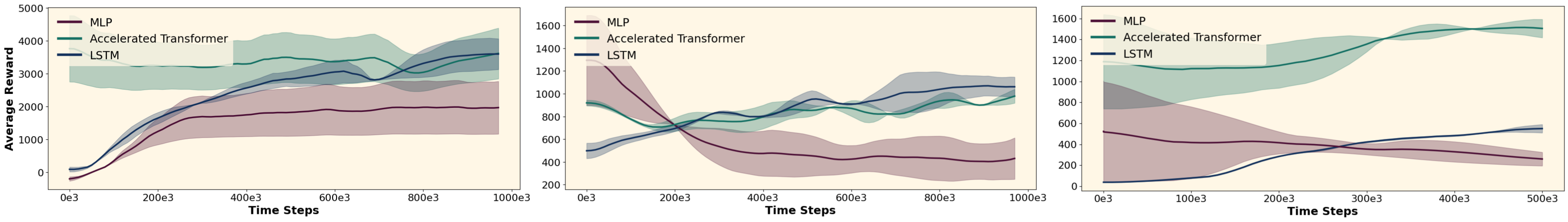}
    \caption{Online training progress of accelerated transformers in the POMDP environments HalfCheetah (left), Ant (center), and Hopper (right). In all three cases, the training algorithm is TD3.}
    \label{fig:rq2_suplement_2}
\end{figure*}

Figure~\ref{fig:rq2_suplement_2} demonstrates that our transformer acceleration enables stable finetuning in the second stage even in POMDP environments. On Hopper (right), the transformer significantly outperforms the LSTM baseline, while on HalfCheetah (left) and Ant (center) its performance is comparable to LSTM. The relation between transformer and baseline results is consistent with the findings of~\citet{resel}, who conducted a more detailed study of POMDP MuJoCo environments. These results indicate that the acceleration algorithm can be generalized to environments with non-Markovian properties, where the accelerator should be a recurrent architecture.

\begin{figure}[!b]
    \centering
    \includegraphics[width=\linewidth]{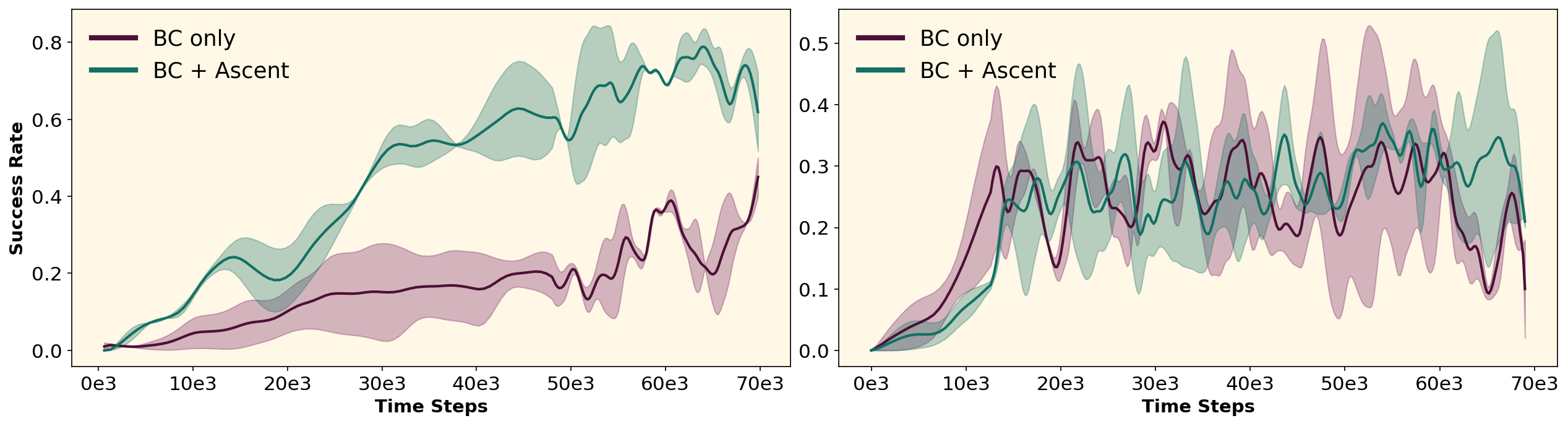}
    \caption{Success rate in PullCube (left) and PushCube (right) environments. Additional gradient ascent is unable to improve or stabilize transformer training during the first stage. }
    \label{fig:bc_vs_asc_main}
\end{figure}

The results of this and the previous section demonstrate the full training pipeline of our method. Although the acceleration stage alone is insufficient to achieve peak performance, this can be compensated by online fine-tuning stage, during which the transformer exhibits stable, high-quality training.
You can find supplementary materials for this experiment in~\autoref{sec: suplement_rq2}.

\paragraph{RQ3. Can additional gradient ascent improve acceleration?}

The Figures in this section present the transformer’s evaluation reward or success rate during the acceleration stage, both with and without the additional ascent. According to Section~\ref{dubsec:one}, there are 
two ways of training the transformer on trajectories from $\mathcal{T}$: by behavior cloning only or with additional gradient ascent over the critic’s function. In this section, we compare these two options to determine whether additional ascent can improve acceleration. To ensure a fair comparison, we use a fixed model and RL parameters listed in~\autoref{tab:opt_params}. In the experiment we investigate whether additional gradient ascent over the accelerator’s critic improves and stabilizes training process. We use the same training data for BC loss calculation and ascending over critic. The frequency of both procedures is the same during the whole training process.

\autoref{fig:bc_vs_asc_main} shows the transformer’s acceleration progress during the first stage, both with and without additional gradient ascent. This technique provides a notable boost in the PullCube task, whereas in the PushCube environment it does not enhance the acceleration process.

Although training with additional ascent sometimes yields better performance, it does not guarantee improvement in every environment. To summarize, the use of additional gradient ascent does not ensure an improvement in the transformer’s training. However, in some cases, it can prove beneficial and provide a performance boost. Each case should be evaluated individually to determine its effectiveness. You can find additional comparisons for this experiment in~\autoref{sec: suplement_rq3}.

\begin{figure}[t]
    \centering
    \includegraphics[width=\linewidth]{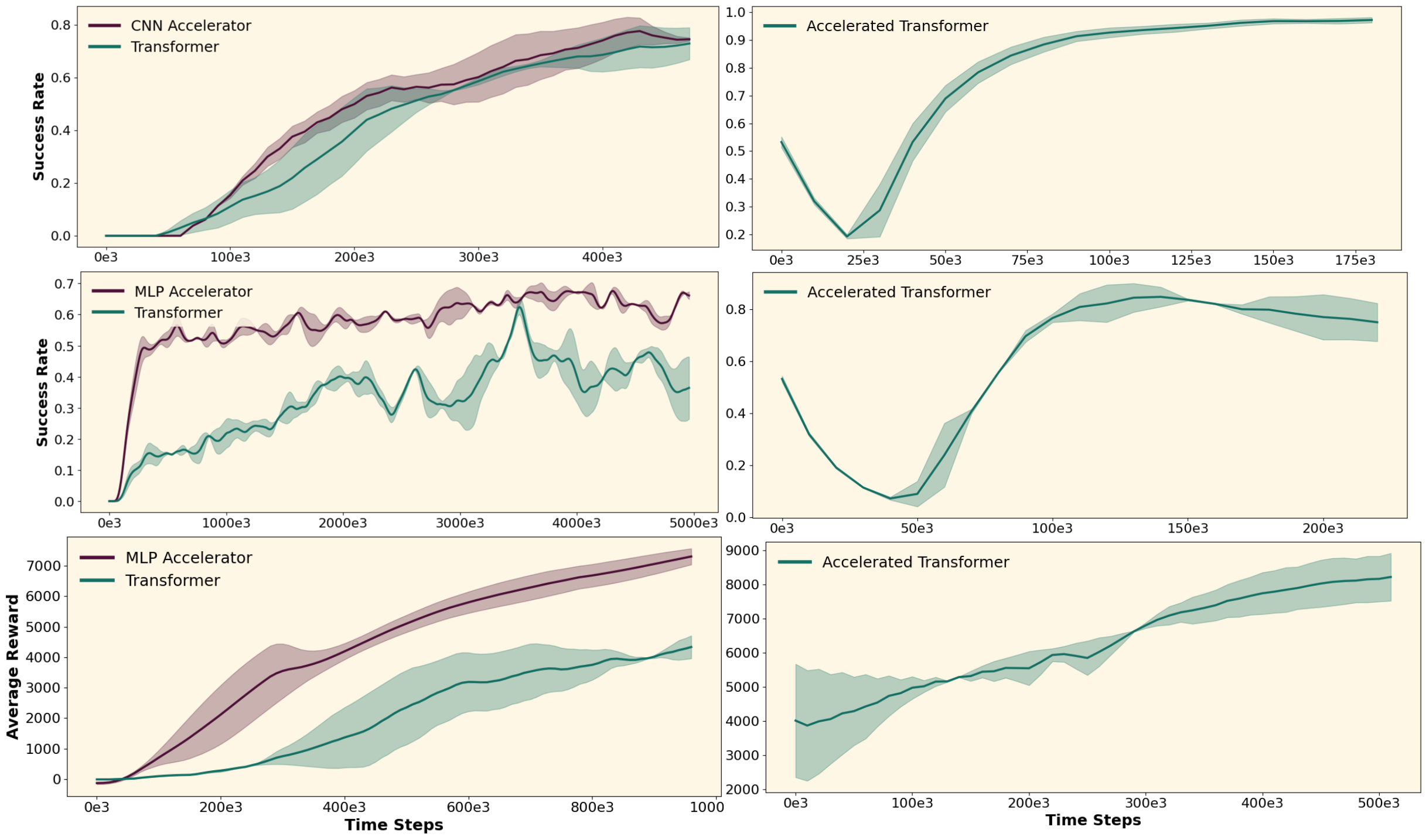}
    \caption{The complete training process on the image-based ManiSkill PullCube task at the top, vector-based PullCube in the middle and on MuJoCo HalfCheetah at the bottom. The left panel corresponds to the first stage of the algorithm, the right panel to the second.}
    \label{fig:rq4_main}
    \vspace{6mm}
\end{figure}

\paragraph{RQ4. Can ``Accelerator" be applied to other off-policy algorithms and image-based environments?}

An important property of a training framework is its ability to generalize across different learning algorithms and to remain effective on more challenging tasks. In this section, we run experiments on the vector-based and image-based versions of PullCube task using Soft Actor-Critic (SAC) algorithm~\citep{sac}, and we additionally conduct an experiment on the HalfCheetah task.
\autoref{fig:rq4_main} presents the results of running the SAC algorithm on the image-based PullCube task (two top plots), the vector-based PullCube task (middle plots), and HalfCheetah (bottom plots). For each environment, the left plot corresponds to the first stage (acceleration) and the right plot to the second stage (online finetuning). In the image-based PullCube task, the CNN accelerator generated trajectories for the transformer, allowing it to pretrain stably and then finetune reliably, achieving a success rate close to 1. It is worth noting that although the vector-based version of the task may appear simpler, training on it turned out to be less stable than on the image-based version.

As can be seen, employing a different off-policy algorithm also enables achieving valuable training performance even in more challenging scenarios such as image-based robotics manipulation tasks.

\paragraph{RQ5. Can ``Accelerator" improve training speed and reduce computational costs?} 

In this section, we highlight two key advantages of our method compared to training from scratch. First, we demonstrate that accelerating the transformer during the first stage substantially reduces the overall training time. This property becomes particularly important in image-based environments with long contexts, where the forward pass is computationally expensive due to quadratic complexity and the backward pass is costly because of convolutional operations.

\begin{figure}
    \centering
\includegraphics[width=1\linewidth]{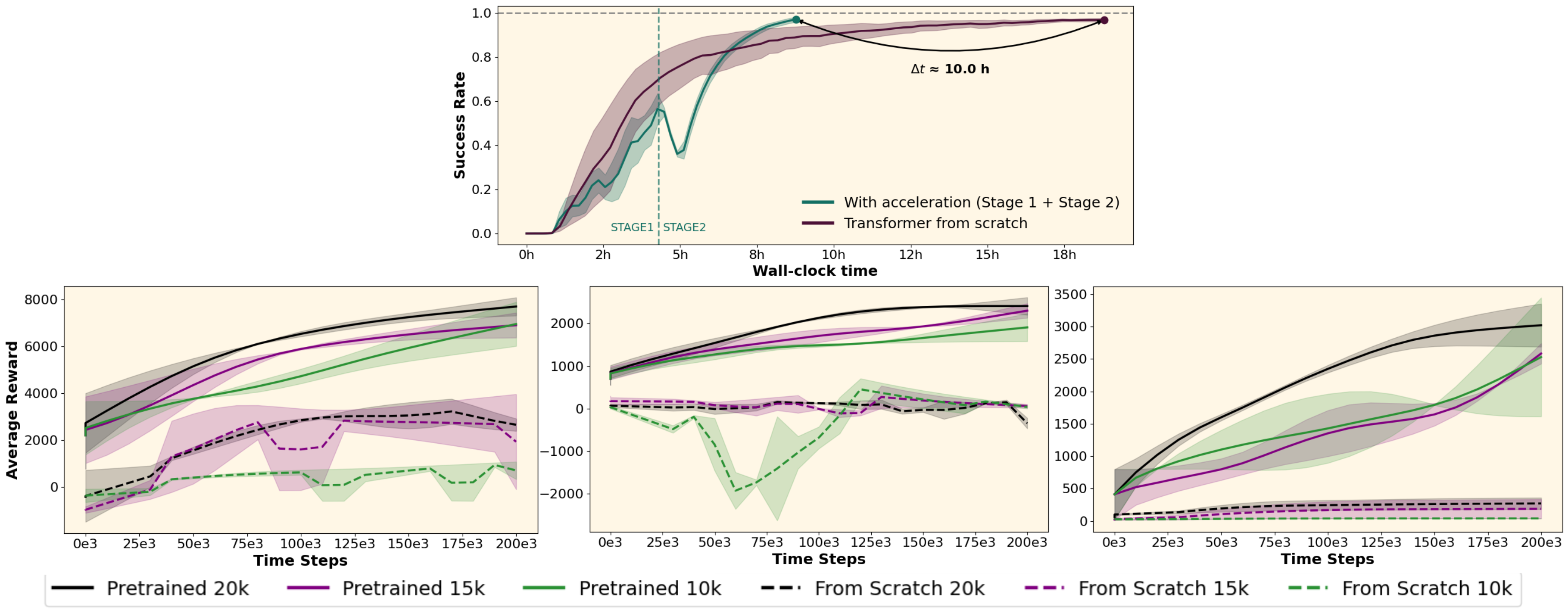}
    \caption{The top plot shows a comparison between training a transformer from scratch and our method in terms of time. It is clearly visible that our approach significantly accelerates training, allowing the model to converge to near-optimal performance on the image-based PullCube task approximately eight hours earlier. 
    The bottom plot shows a comparison of the accelerated and non-accelerated transformer training performance with $\mathcal{B}$ of shape 10, 15 and 20 thousands. HalfCheetah (left), Ant (middle) and Hopper (right) environments.}
    \label{fig:rb_variation}
    \vspace{7mm}    
\end{figure}

For illustration, the top part of~\autoref{fig:rb_variation} compares the training duration of a transformer trained from scratch against our method in image-based PullCube task with context of 3. The green dashed line marks the transition from the first stage to the second stage of training. This division applies only to the green curve, since the red curve (from scratch) was trained exclusively in the online setting. It is clearly visible that BC in the first stage makes the transformer more robust and provides greater potential for rapid learning during the online finetuning stage. The final difference in convergence speed to near-optimal success rate values amounted to roughly 10 hours.

In addition to improving training speed, it is also desirable to reduce the memory cost of training the transformer. To address this question, it is necessary to consider the design of off-policy algorithms, which use a replay buffer for accumulating experience during the agent’s training. To maintain high-quality training and the necessary level of exploration, it is essential to create large replay buffers, typically on the order of 1{,}000{,}000 observations~\citep{td3}. Training the transformer on multiple environments in parallel, along with a large context, may require significant memory to store a large replay buffer. In image-based environments, the allocated memory increases even more due to the need to store images instead of vectors.
Thus, while online training from scratch is feasible for some environments, the proposed offline pre-training approach offers a potential solution to manage memory constraints and improve training efficiency, particularly in scenarios involving large-scale data and parallel environment interactions.

In our algorithm, the trajectory buffer $\mathcal{T}$ only needs to store fresh training data corresponding to the accelerator’s current skills, eliminating the need to initialize $\mathcal{T}$ with a large size. As a result, throughout the entire acceleration stage, our memory costs for storing the trajectory buffer remain minimal.
Furthermore, during the online fine-tuning stage, the pre-trained transformer also does not require a large replay buffer. The need for extensive environment exploration has already been satisfied during the acceleration stage, so the replay buffer $\mathcal{B}$ does not need to store highly diverse experience. Consequently, the fine-tuning process can proceed efficiently with reduced memory overhead, making the algorithm well-suited for resource-constrained systems.

In the experiment shown in~\autoref{fig:rb_variation}, we ran both the accelerated and non-accelerated transformers with varying replay buffer sizes to highlight a key advantage of our algorithm. We selected replay buffer sizes of 10{,}000, 15{,}000 and 20{,}000 and trained each model with these sizes.
The results clearly show that the non-accelerated version failed to train effectively with any of the smaller buffer sizes.
In contrast, our model demonstrated effective training with small buffer sizes and only began to slightly lose quality at the size of 15{,}000 and 10{,}000 in the Ant environment. These findings indicate that the effective threshold for the replay buffer size lies between 20{,}000 and 10{,}000, which is more than 50 times smaller than 1{,}000{,}000, the standard size of the buffer.

\section{Discussion and future work}
We proposed a two-stage algorithm for accelerating transformers in online reinforcement learning, addressing key challenges of fully online and offline pretraining such as unstable training, limited datasets, weak exploration, and state distribution shifts. By combining online data collection with stable pretraining and subsequent fine-tuning, our method achieves performance comparable to MLP and LSTM baselines, while also reducing computational costs and training time. These results highlight the importance of transformer acceleration as a promising direction for making transformer-based RL more stable and accessible. Among the current limitations of our approach is the absence of an automatic switching mechanism upon reaching a formal condition, which requires continuous involvement and monitoring of the transformer’s acceleration process until its progress during the pretraining phase saturates.
We hope that our work will inspire further advances in this direction.

\clearpage
\bibliography{main}

\clearpage
\onecolumn
\appendix
\section{Supplementary materials for experimental setup}
\label{appendix}

In~\autoref{tab:opt_params},~\autoref{tab:opt_params_sac} and~\autoref{tab:opt_params_lstm} the first two columns describe MLP accelerator's parameters that were used during acceleration stage with TD3 and SAC respectively. The second two columns describe transformers parameters that were used during fine-tune stage. Since model parameters are fixed during both stages, this table also consists of all transformer's parameters that were used in acceleration stage too.

\begin{table}[th]
\caption{Parameters used for all the experiments with TD3 algorithm.}
\label{tab:opt_params}
\vskip 0.15in
\begin{center}
\begin{small}
\begin{sc}
\begin{tabular}{lccccc}
\toprule
 & \multicolumn{2}{c}{1st stage accelerator params} & \multicolumn{2}{c}{1st and 2nd stage transformer params} \\
\cmidrule(lr){2-3} \cmidrule(lr){4-5}
Parameter & ManiSkill & MuJoCo & ManiSkill & MuJoCo \\
\midrule
$\gamma$-discount  & 0.8 & 0.99 & 0.8 & 0.99 \\
$\tau$-soft update & 0.01 & 0.005 & 0.01 & 0.005 \\
Policy noise       & 0.2 & 0.2 & 0.2 & 0.2 \\
Noise clip         & 0.5 & 0.5 & 0.5 & 0.5 \\
Exploration noise  & 0.1 & 0.1 & 0.1 & 0.1 \\
Batch size         & 600 & 256 & 256 & 256 \\
Optimizer          & Adam & Adam & Adam & Adam \\
Learning Rate      & $3\times10^{-4}$ & $3\times10^{-4}$ & $3\times10^{-4}$ & $3\times10^{-4}$ \\
Buffer size        & $0.05\times10^{6}$ & $0.5\times10^{6}$ & $0.01\times10^{6}$ & $0.1\times10^{6}$\\
Learning Starts    & 600 & 25000 & 0 & 0 \\
Num Envs    & 50 & 1 & 50 & 1 \\
\midrule
Num layers        & 2   & 2    & 1      & 1 \\
Num heads         & -   & -    & 2      & 2 \\
Dim model         & -   & -    & 256    & 256 \\
Dim feedforward   & 256 & 256  & 512    & 512 \\
Dropout           & -   & -    & 0.05   & 0.05 \\
Context len       & -   & -    & 3      &    3 \\
\bottomrule
\end{tabular}
\end{sc}
\end{small}
\end{center}
\vskip -0.1in
\end{table}

\begin{table}[th]
\caption{Parameters used for all the experiments with SAC algorithm.}
\label{tab:opt_params_sac}
\vskip 0.15in
\begin{center}
\begin{small}
\begin{sc}
\begin{tabular}{lccccc}
\toprule
 & \multicolumn{2}{c}{1st stage accelerator params} & \multicolumn{2}{c}{1st and 2nd stage transformer params} \\
\cmidrule(lr){2-3} \cmidrule(lr){4-5}
Parameter & ManiSkill & MuJoCo & ManiSkill & MuJoCo \\
\midrule
$\gamma$-discount     & 0.99 & 0.99 & 0.99 & 0.99 \\
$\tau$-soft update    & 0.005 & 0.007 & 0.005 & 0.007 \\
Exploration noise     & 0.1 & 0.1 & 0.1 & 0.1 \\
Batch size            & 256 & 256 & 256 & 256 \\
Optimizer             & Adam & Adam & Adam & Adam \\
Actor LR              & $3\times10^{-4}$ & $3\times10^{-4}$ & $3\times10^{-4}$ & $3\times10^{-4}$ \\
Critic LR             & $10^{-3}$ & $10^{-3}$ & $10^{-3}$ & $10^{-3}$ \\
$\alpha$ (entropy)    & learnable & learnable & learnable & learnable \\
$\alpha$ LR           & $3\times10^{-4}$ & $3\times10^{-4}$ & $3\times10^{-4}$ & $3\times10^{-4}$ \\
Buffer size           & $0.05\times10^{6}$ & $1.0\times10^{6}$ & $0.03\times10^{6}$ & $1.0\times10^{6}$ \\
Learning Starts       & 5000 & 1000 & 5000 & 1000 \\
Num Envs              & 50 & 1 & 50 & 1 \\
\midrule
Num layers            & 2   & 2    & 2      & 2 \\
Num heads             & -   & -    & 2      & 2 \\
Dim model             & -   & -    & 256    & 256 \\
Dim feedforward       & 256 & 256  & 512    & 512 \\
Dropout               & -   & -    & 0.05   & 0.05 \\
Context len           & -   & -    & 5      & 5 \\
\bottomrule
\end{tabular}
\end{sc}
\end{small}
\end{center}
\vskip -0.1in
\end{table}

\begin{table}[th]
\caption{Parameters used for all the experiments with TD3 algorithm and LSTM accelerator.}
\label{tab:opt_params_lstm}
\vskip 0.15in
\begin{center}
\begin{small}
\begin{sc}
\begin{tabular}{lcc}
\toprule
 & Accelerator (stage 1) & Transformer (stages 1 \& 2) \\
\midrule
$\gamma$-discount     & 0.99 & 0.99 \\
$\tau$-soft update    & 0.007 & 0.007 \\
Policy noise          & 0.2 & 0.2 \\
Noise clip            & 0.2 & 0.2 \\
Exploration noise     & 0.1 & 0.1 \\
Batch size            & 256 & 256 \\
Optimizer             & Adam & Adam \\
Learning rate         & $3\times 10^{-4}$ & $3\times 10^{-4}$ \\
Buffer size           & $1.0\times 10^{6}$ & $0.03\times 10^{6}$ \\
Learning starts       & 0 & 0 \\
Num envs              & 1 & 1 \\
Grad clip             & $1\times 10^{5}$ & $1\times 10^{5}$ \\
\midrule
Num layers            & 2 (LSTM) & 1 (MLP) \\
Num heads             & 2 & 2 \\
Dim model             & 256 & 256 \\
Dim feedforward       & 256 & 512 \\
Dropout               & 0.05 & 0.05 \\
Context length        & 5 & 5 \\
\bottomrule
\end{tabular}
\end{sc}
\end{small}
\end{center}
\vskip -0.1in
\end{table}

\section{Supplementary materials for RQ1}
\label{sec: suplement_rq1}

\begin{figure}[H]
    \centering
    \includegraphics[width=0.85\linewidth]{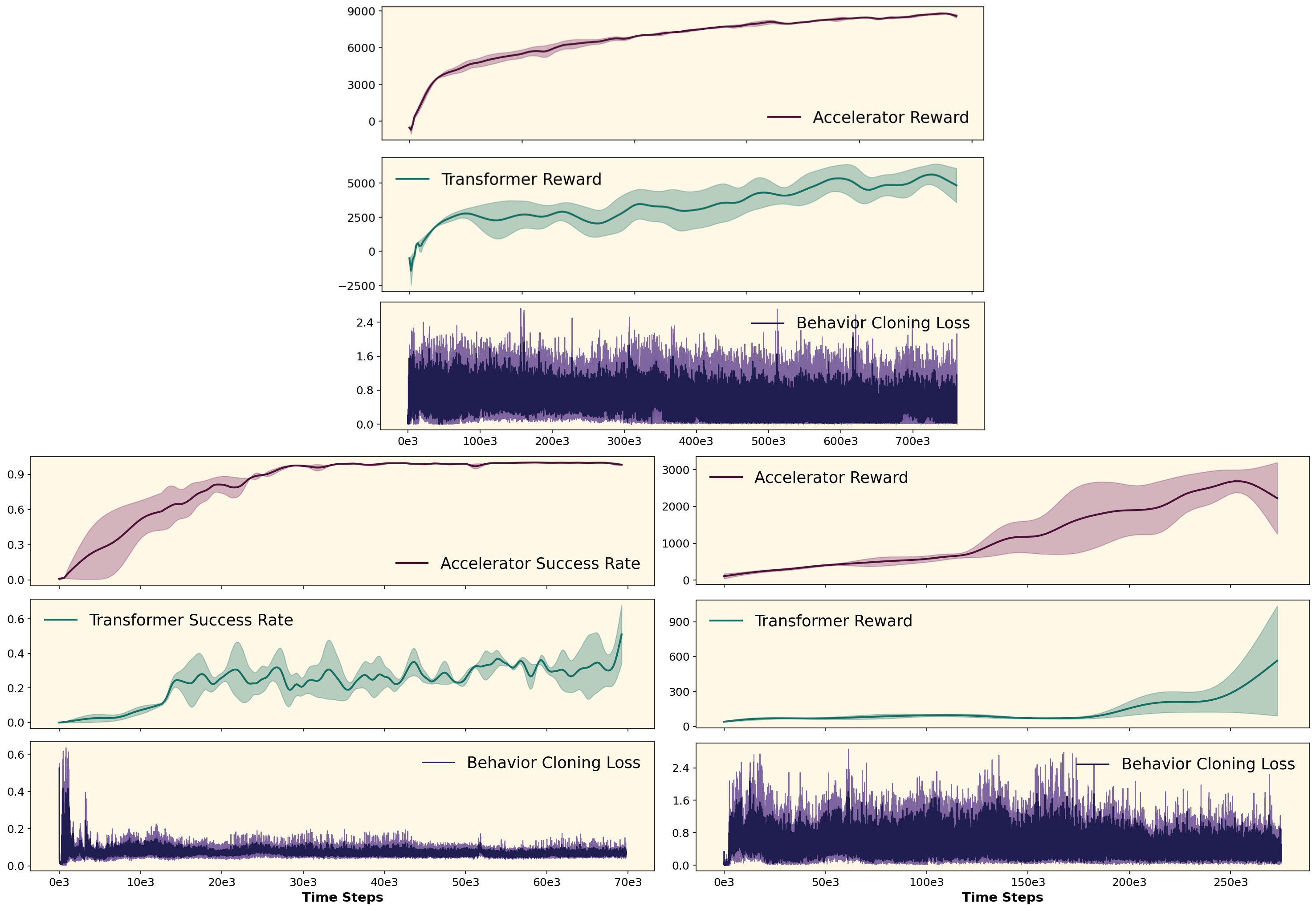}
    \caption{PushCube (bottom left), Hopper (bottom right) and HalfCheetah (top) acceleration progress.}
    \label{fig:st_1_appendix}
\end{figure}

\begin{figure}[H]
    \centering
    \includegraphics[width=0.85\linewidth]{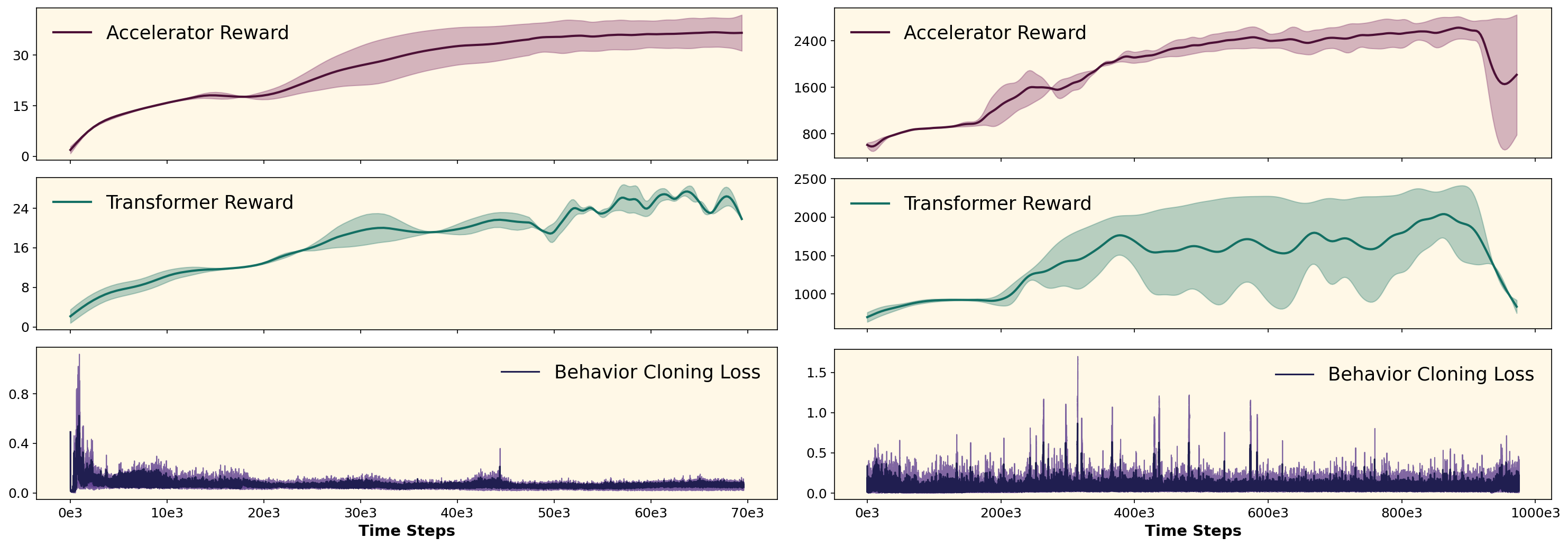}
    \caption{PullCube (left), Ant (right) acceleration progress.}
    \label{fig:st_2_appendix}
\end{figure}

Figures~\ref{fig:st_1_appendix} and~\ref{fig:st_2_appendix} illustrate the acceleration process of the transformer on the MDP versions of the MuJoCo and ManiSkill environments. In all experiments, the TD3 algorithm was used. It can be observed that the transformer’s progress largely mirrors that of the accelerator, represented by the MLP. For instance, an unexpected performance drop of the accelerator in Ant is almost immediately followed by a similar drop in the transformer’s performance.

\vspace{1mm}
\section{Supplementary materials for RQ2}
\label{sec: suplement_rq2}
\begin{figure}[H]
    \centering
    \includegraphics[width=0.7\linewidth]{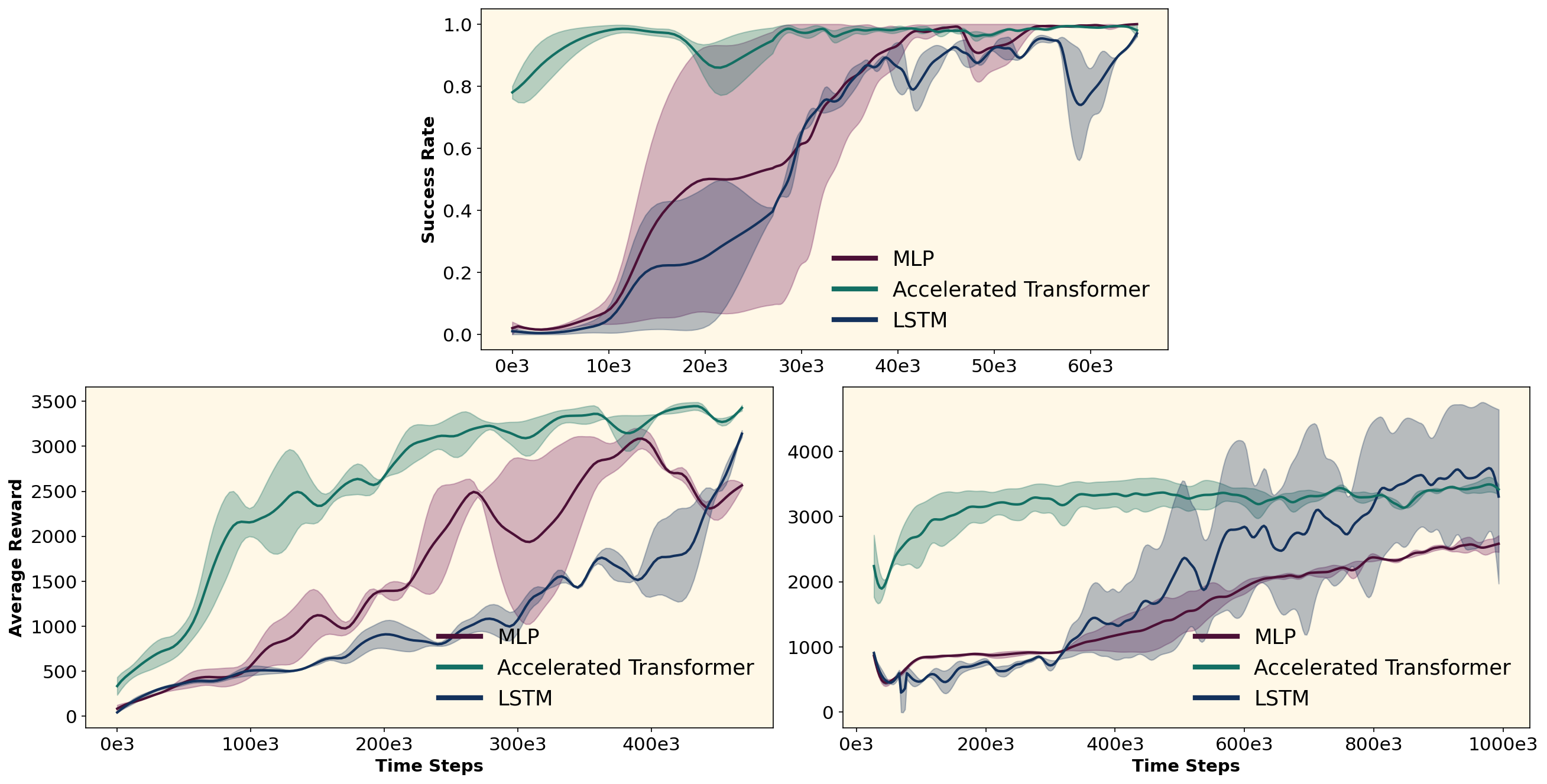}
    \caption{Online training progress. Success rate in PullCube (top) and average reward on Hopper (bottom left) and Ant (bottom right) tasks.}
    \label{fig:rq2_suplement_1}
\end{figure}

As shown in Figure~\ref{fig:rq2_suplement_1}, in MDP environments the accelerated transformer achieves performance comparable to, and in some cases exceeding, the MLP and LSTM baselines.

\section{Supplementary materials for RQ3}
\label{sec: suplement_rq3}
\begin{figure}[H]
    \centering
    \includegraphics[width=1\linewidth]{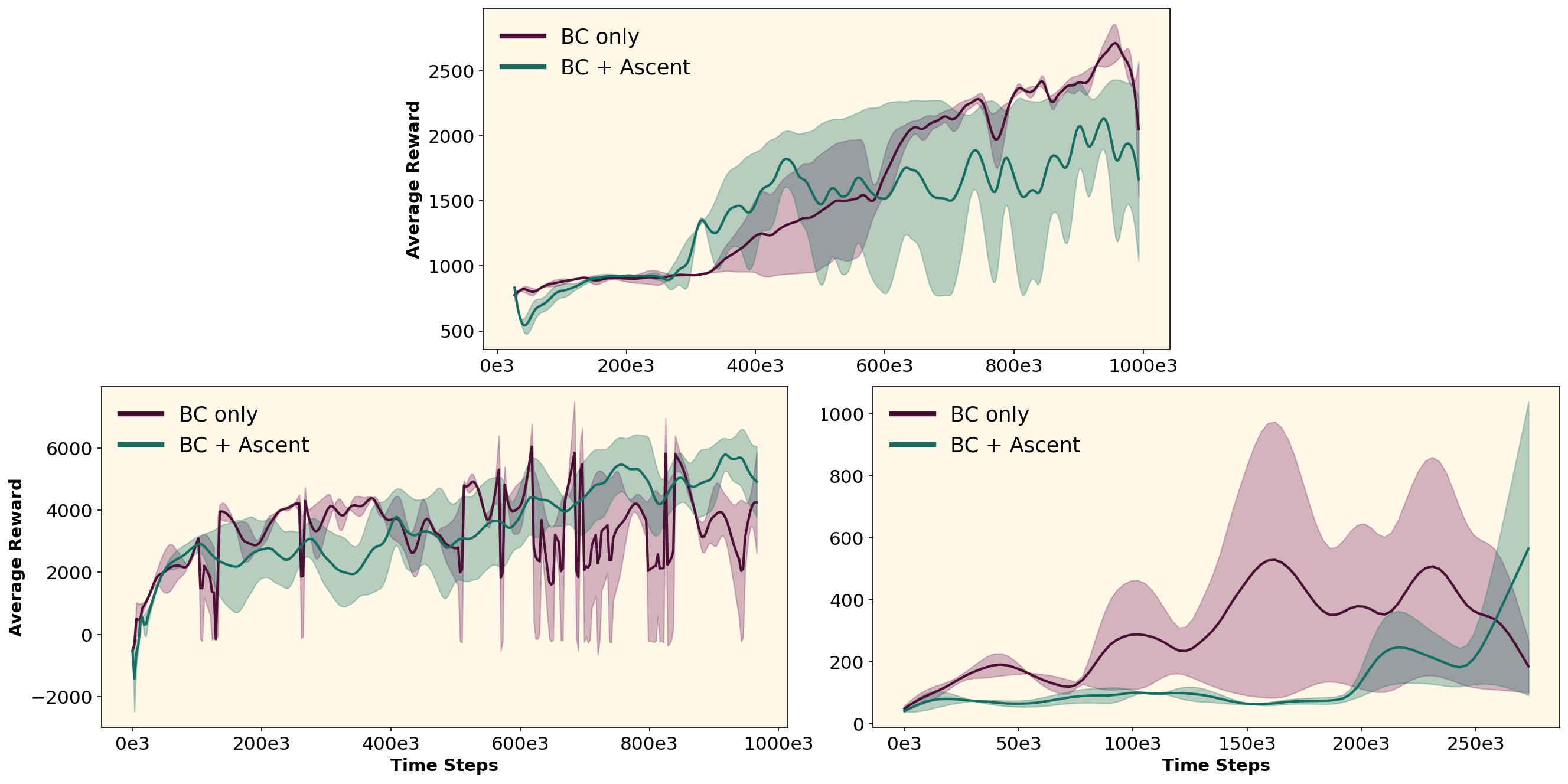}
    \caption{Acceleration progress with and without additional ascent. Ant (top), HalfCheetah (bottom left) and Hopper (bottom right) environments.}
    \label{fig:bc_vs_asc_appendix}
\end{figure}

The experiments did not reveal a clear advantage of additional gradient ascent, as evidenced by Figure~\ref{fig:bc_vs_asc_appendix}.

\end{document}